\documentclass{ieeeaccess}
\usepackage{cite}
\usepackage{amsmath,amssymb,amsfonts}
\usepackage{algorithmic}
\usepackage{graphicx}
\usepackage{textcomp}
\usepackage{subfig}
\usepackage{hyperref}
\usepackage{fontawesome}

\usepackage{colortbl,dcolumn}
\definecolor{red}{rgb}{0.9,0.2,0}
\definecolor{teal}{rgb}{0.18,0.49,0.20}
\definecolor{green}{rgb}{0.29,0.69,0.32}
\definecolor{greenlight}{rgb}{0.55,0.77,0.29}
\definecolor{lime}{rgb}{0.80,0.86,0.22}
\definecolor{yellow1}{rgb}{1,0.92,0.23}
\definecolor{amber}{rgb}{1,0.76,0.03}
\definecolor{orange}{rgb}{1,0.6,0}
\definecolor{deeporange}{rgb}{1,0.34,0.14}

\definecolor{mygray}{rgb}{0.70,0.70,0.70}

\def\zz#1{%
\ifdim#1pt<1pt\cellcolor{teal}\else
\ifdim#1pt<4pt\cellcolor{green}\else
\ifdim#1pt<8pt\cellcolor{greenlight}\else
\ifdim#1pt<16pt\cellcolor{lime}\else
\ifdim#1pt<33pt\cellcolor{yellow1}\else
\ifdim#1pt<66pt\cellcolor{amber}\else
\ifdim#1pt<200pt\cellcolor{orange}\else
\cellcolor{deeporange}\fi\fi\fi\fi\fi\fi\fi
#1}

\def\mem#1{%
\ifdim#1pt<1pt\cellcolor{teal}\else
\ifdim#1pt<2pt\cellcolor{green}\else
\ifdim#1pt<3pt\cellcolor{greenlight}\else
\ifdim#1pt<4pt\cellcolor{lime}\else
\ifdim#1pt<5pt\cellcolor{yellow1}\else
\ifdim#1pt<6pt\cellcolor{amber}\else
\ifdim#1pt<7pt\cellcolor{orange}\else
\cellcolor{deeporange}\fi\fi\fi\fi\fi\fi\fi
#1}

\def\BibTeX{{\rm B\kern-.05em{\sc i\kern-.025em b}\kern-.08em
    T\kern-.1667em\lower.7ex\hbox{E}\kern-.125emX}}
\begin{document}
\history{Date of publication xxxx 00, 0000, date of current version xxxx 00, 0000.}
\doi{10.1109/ACCESS.2017.DOI}

\title{Benchmark Analysis of Representative\\Deep Neural Network Architectures}
\author{\uppercase{Simone Bianco\authorrefmark{1}}, \uppercase{Remi Cadene\authorrefmark{2}}, \uppercase{Luigi Celona\authorrefmark{1}, and Paolo Napoletano\authorrefmark{1}}.
}
\address[1]{University of Milano-Bicocca, Department of Informatics, Systems and Communication, viale Sarca, 336, 20126 Milano, Italy \\
}
\address[2]{Sorbonne Universit\'e, CNRS, LIP6, F-75005 Paris, France %(e-mail: remi.cadene@lip6.fr)
}

\pubid{\copyright 2018 IEEE}

%\tfootnote{This paragraph of the first footnote will contain support information, including sponsor and financial support acknowledgment. For example, ``This work was supported in part by the U.S. Department of Commerce under Grant BS123456.''}

\markboth
{Author \headeretal: Preparation of Papers for IEEE TRANSACTIONS and JOURNALS}
{Author \headeretal: Preparation of Papers for IEEE TRANSACTIONS and JOURNALS}

\corresp{Corresponding author: Luigi Celona (e-mail: luigi.celona@disco.unimib.it).\\
\faGithub $\ $ \url{https://github.com/CeLuigi/models-comparison.pytorch}}

\begin{abstract}
This work presents an in-depth analysis of the majority of the deep neural networks (DNNs) proposed in the state of the art for image recognition. For each DNN multiple performance indices are observed, such as recognition accuracy, model complexity, computational complexity, memory usage, and inference time. The behavior of such  performance indices and some combinations of them are analyzed and discussed. 
To measure the indices we experiment the use of DNNs on two different computer architectures, a workstation equipped with a NVIDIA Titan X Pascal and an embedded system based on a NVIDIA Jetson TX1 board. This experimentation allows a direct comparison between DNNs running on machines with very different computational capacity. This study is useful for researchers to have a complete view of what solutions have been explored so far and in which research directions are worth exploring in the future; and for practitioners to select the DNN architecture(s) that better fit the resource constraints of practical deployments and applications. To complete this work, all the DNNs, as well as the software used for the analysis, are available online.
\end{abstract}
\begin{keywords}
Deep neural networks, Convolutional neural networks, Image recognition.
%Enter key words or phrases in alphabetical order, separated by commas. For a list of suggested keywords, send a blank e-mail to keywords@ieee.org or visit \underline {http://www.ieee.org/organizations/pubs/ani\_prod/keywrd98.txt}
\end{keywords}

\titlepgskip=-15pt

\maketitle
\section{Introduction}
\label{sec:introduction}
Deep neural networks (DNNs) have achieved remarkable results in many computer vision tasks~\cite{lecun2015deep}. AlexNet~\cite {krizhevsky2012imagenet}, that is the first DNN presented in the literature in 2012, drastically increased the recognition accuracy (about 10\% higher) with respect to traditional methods on the 1000-class ImageNet Large-Scale Visual Recognition Competition (ImageNet-1k) \cite{russakovsky2015imagenet}. Since then, literature has worked both in designing more accurate networks as well as in designing more efficient networks from a computational-cost point of view.

Although there is a lot of literature discussing new architectures from the point of view of the layers composition and recognition performance, there are few papers that analyze the aspects related to the computational cost (memory usage, inference time, etc.), and more importantly how  computational cost impacts on the recognition accuracy. 

Canziani \emph{et al.} \cite{canziani2016analysis} in the first half of 2016 proposed a comprehensive analysis of some DNN architectures by performing experiments on an embedded system based on a NVIDIA Jetson TX1 board. They measured accuracy, power consumption, memory footprint, number of parameters and operations count, and more importantly they analyzed the relationship between these performance indices. It is a valuable work, but it has been focused on a limited number (i.e. 14) of DNNs and more importantly the experimentation has been carried out only on the NVIDIA Jetson TX1 board. In \cite{huang2017speed}, speed/accuracy trade-off of modern DNN-based detection systems has been explored by re-implementing a set of meta-architectures inspired by well-known detection networks in the state of the art. Results include performance comparisons between an Intel Xeon CPU and a NVIDIA Titan X GPU.

The aim of this work is to provide a more comprehensive and complete analysis of existing DNNs for image recognition and most importantly to provide an analysis on two hardware platforms with a very different computational capacity: a workstation equipped with a NVIDIA Titan X Pascal (often referred to as Titan Xp) and an embedded system based on a NVIDIA Jetson TX1. To this aim we analyze and compare more than 40 state-of-the-art DNN architectures in terms of computational cost and accuracy. In particular we experiment the selected DNN architectures on the ImageNet-1k challenge and we measure: accuracy rate, model complexity, memory usage, computational complexity, and inference time. Further, we analyze relationships between these performance indices that provide insights for: 1) understanding what solutions have been explored so far and in what direction it would be appropriate to go in the future; 2) selecting the DNN architecture that better fits the resource constraints of practical deployments and applications.

The most important findings are that: 
i) the recognition accuracy does not increase as the number of operations increases;
ii) there is not a linear relationship between model complexity and accuracy;
iii) the desired throughput places an upper bound to the achievable accuracy;
iv) not all the DNN models use their parameters with the same level of efficiency;
v) almost all models are capable of super real-time performance on a high-end GPU, while just some of them can guarantee it on an embedded system;
vi) even  DNNs  with  a  very  low  level  model  complexity have a minimum GPU memory footprint of about 0.6GB.

The rest of paper is organized as follows: in Section \ref{sec:benchmarking} hardware and software used for experiments are detailed; in Section \ref{sec:architectures} the considered DNN architectures are briefly introduced; in Section \ref{sec:metrics} the measured performance indices are described; finally, in Section \ref{sec:results} obtained results are reported and analyzed, and Section \ref{sec:discussion} presents our final considerations.
\section{Benchmarking}
\label{sec:benchmarking}
We implement the benchmark framework for DNNs comparison in Python. The PyTorch package \cite{paszke2017automatic} is used for neural networks processing with cuDNN-v5.1 and CUDA-v9.0 as back-end.  All the code for the estimation of the adopted performance indices, as well as all the considered DNN models are made publicly available~\cite{github_celona}.%\footnote{\faGithub $\ $ \url{https://github.com/CeLuigi/models-comparison.pytorch}}

We run all the experiments on a workstation and on an embedded system: 
\begin{enumerate}
    \item The workstation is equipped with an Intel Core I7-7700 CPU @ 3.60GHZ, 16GB DDR4 RAM 2400 MHz, NVIDIA Titan X Pascal GPU with 3840 CUDA cores (top-of-the-line consumer GPU). The operating system is Ubuntu 16.04. 
    \item The embedded system is a NVIDIA Jetson TX1 board with 64-bit ARM\textsuperscript \textregistered A57 CPU @ 2GHz, 4GB LPDDR4 1600MHz, NVIDIA Maxwell GPU with 256 CUDA cores. The board includes the JetPack-2.3 SDK.
\end{enumerate} 
The use of these two different systems allows to highlight how critical the computational resources can be depending on the DNN model adopted especially in terms of memory usage and inference time. 
\section{Architectures}
\label{sec:architectures}
In this section we briefly describe the analyzed architectures. We select different architectures, some of which have been designed to be more performing in terms of effectiveness, while others have been designed to be more efficient and therefore more suitable for embedded vision applications. In some cases there is a number following the name of the architecture. Such a number depicts the number of layers that contains parameters to be learned (i.e. convolutional or fully connected layers).

We consider the following architectures: AlexNet \cite{krizhevsky2012imagenet}; the family of VGG architectures \cite{simonyan2014very} (VGG-11, -13, -16, and -19) without and with the use of Batch Normalization (BN) layers \cite{ioffe2015batch}; BN-Inception \cite{ioffe2015batch}; GoogLeNet \cite{Szegedy_2015_CVPR}; SqueezeNet-v1.0 and -v1.1 \cite{iandola2016squeezenet}; ResNet-18, -34, -50, -101, and -152 \cite{he2016deep}; Inception-v3 \cite{szegedy2016rethinking}; Inception-v4 and Inception-ResNet-v2 \cite{szegedy2016inc}; DenseNet-121, -169, and -201 with growth rate corresponding to 32, and DenseNet-161 with growth rate equal to 48 \cite{huang2017densely}; ResNeXt-101 (32x4d), and ResNeXt-101 (64x4d), where the numbers inside the brackets denote respectively the number of groups per convolutional layer and the bottleneck width \cite{xie2017aggregated}; Xception \cite{Chollet_2017_CVPR}; DualPathNet-68, -98, and -131, \cite{chen2017dual}; SE-ResNet-50, SENet-154, SE-ResNet-101, SE-ResNet-152, SE-ResNeXt-50 (32x4d), SE-ResNeXt-101 (32x4d) \cite{Hu_2018_CVPR}; NASNet-A-Large, and NASNet-A-Mobile, whose architecture is directly learned \cite{Zoph_2018_CVPR}.

Furthermore, we also consider the following efficientcy-oriented models: MobileNet-v1 \cite{howard2017mobilenets}, MobileNet-v2 \cite{sandler2018mobilenetv2}, and ShuffleNet \cite{Zhang_2018_CVPR}.
\section{Performance indices}
\label{sec:metrics}
In order to perform a direct and fair comparison, we exactly reproduce the same sampling policies: we directly collect models trained using the PyTorch framework \cite{paszke2017automatic}, or we collect models trained with other deep learning frameworks and then we convert them in PyTorch.

All the pre-trained models expect input images normalized in the same way, i.e. mini-batches of RGB images with shape $3 \times H \times W$, where $H$ and $W$ are expected to be: 
\begin{itemize}
    \item[-] 331 pixels for the NASNet-A-Large model;
    \item[-] 229 pixels for InceptionResNet-v2, Inception-v3, Inception-v4, and Xception models;
    \item[-] 224 pixels for all the other models considered.
\end{itemize}  

We consider multiple performance indices useful for a comprehensive benchmark of DNN models. Specifically, we measure: accuracy rate, model complexity, memory usage, computational complexity, and inference time.
\subsection{Accuracy rate}
\label{sec:accuracy}
We estimate Top-1 and Top-5 accuracy on the ImageNet-1k validation set for image classification task. The predictions are computed by evaluating the central crop only. Slightly better performances can be achieved by considering the average prediction coming from multiple crops (four corners plus central crop and their horizontal flips).
\subsection{Model complexity}
\label{sec:model_complexity}
We analyze model complexity by counting the total amount of learnable parameters. Specifically, we collect the size of the parameter file in terms of MB for the considered models. This information is very useful for understanding the minimum amount of GPU memory required for each model.

\begin{figure*}[!ht]
\centering
\resizebox{\textwidth}{!}{
    \begin{tabular}{cc}
    \includegraphics[height=8truecm]{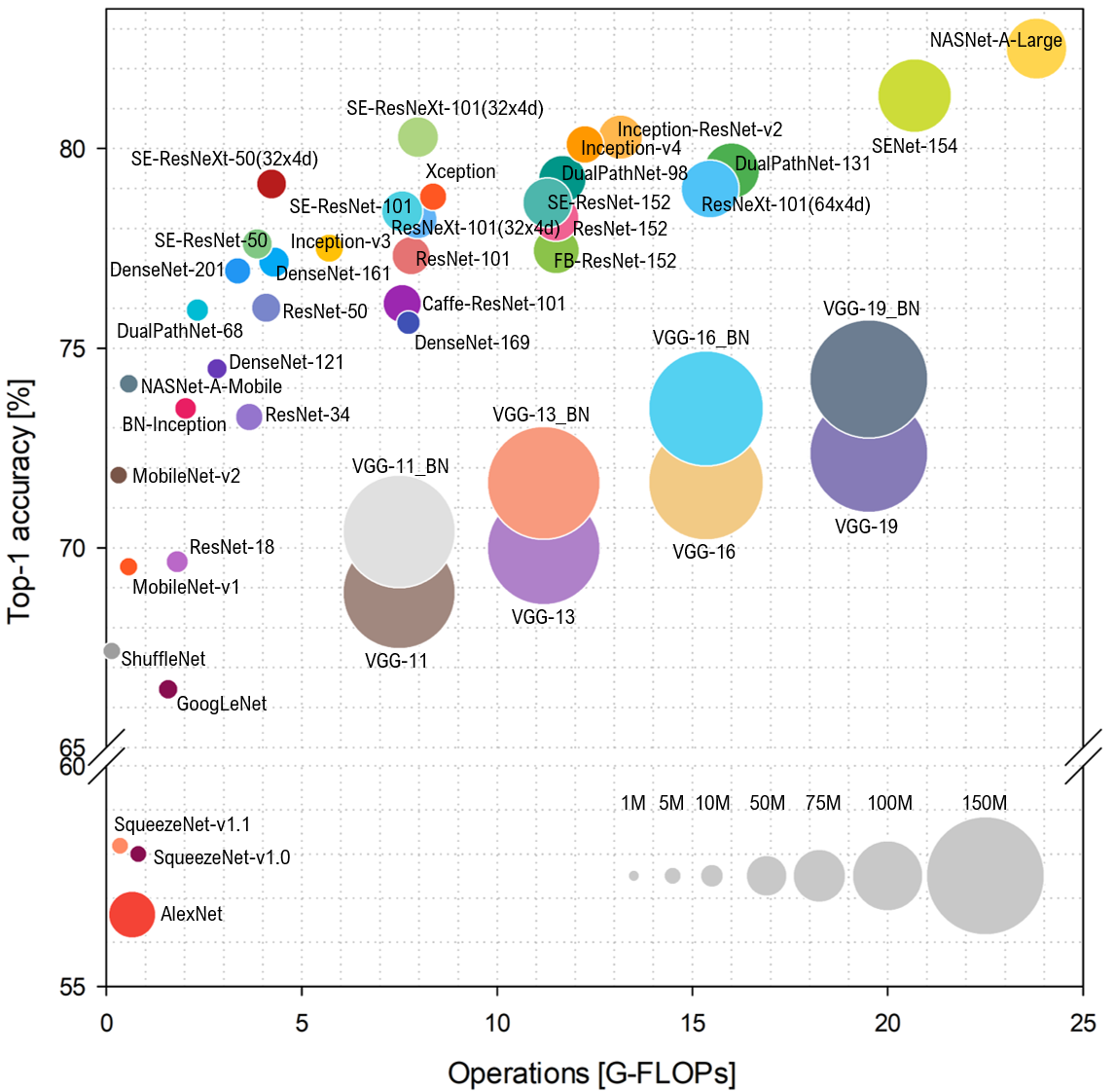} &
    \includegraphics[height=8truecm]{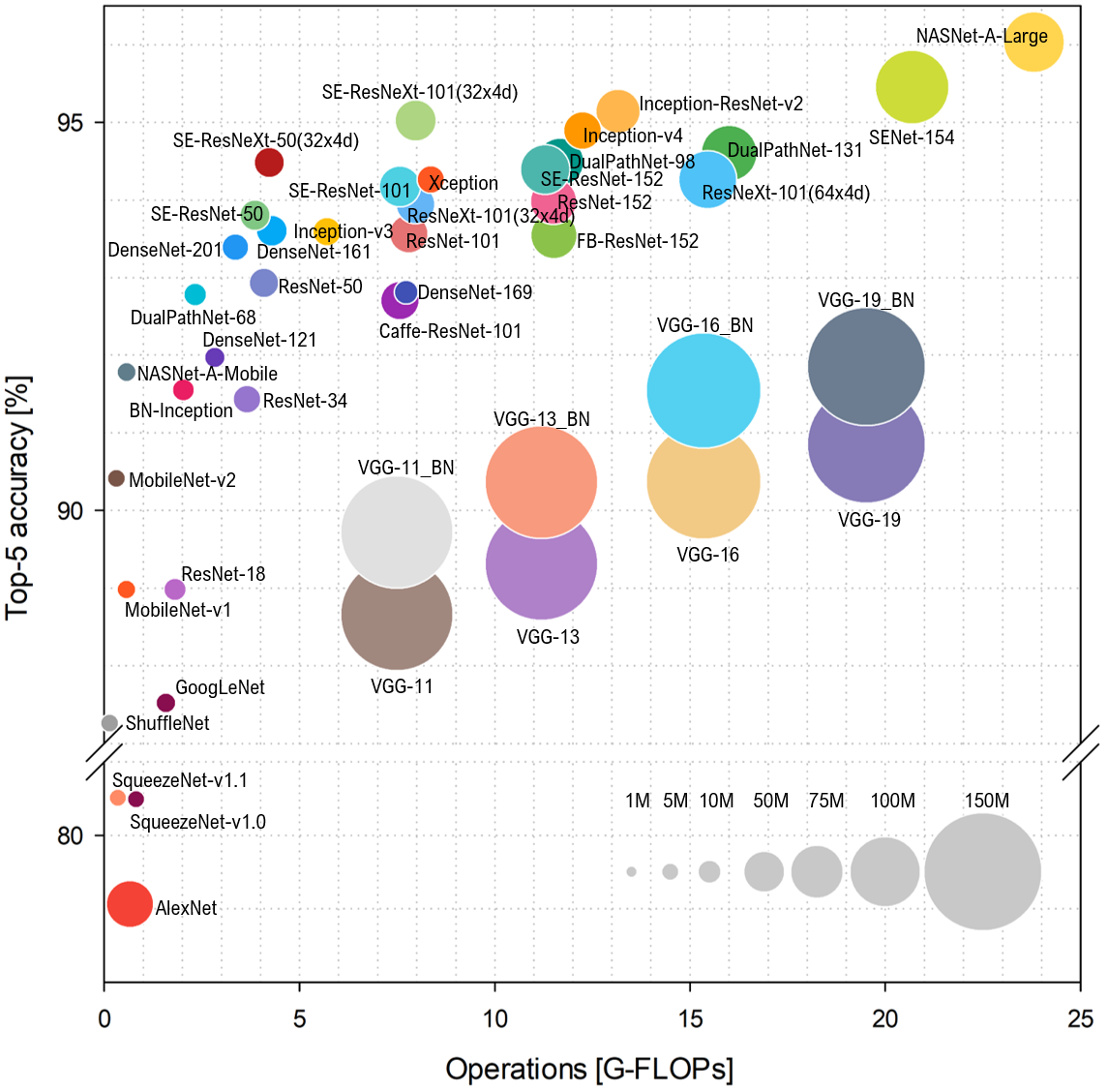} \\
    (a)&(b)\\
    \end{tabular}}
    \caption{Ball chart reporting the Top-1 and Top-5 accuracy \emph{vs.} computational complexity. Top-1 and Top-5 accuracy using only the center crop versus floating-point operations (FLOPs) required for a single forward pass are reported. The size of each ball corresponds to the model complexity. (a) Top-1; (b) Top-5.}
    \label{fig:accuracyvsflops}
\end{figure*}

\subsection{Memory usage}
\label{sec:memory}

We evaluate the total memory consumption, which includes all the memory that is allocated, i.e. the memory allocated for the network model and the memory required while processing the batch. We measure memory usage for different batch sizes: 1, 2, 4, 8, 16, 32, and 64.
\subsection{Computational complexity}
\label{sec:computational_complexity}
We measure the computational cost of each DNN model considered using the floating-point operations (FLOPs) in the number of multiply-adds as in \cite{xie2017aggregated}. More in detail, the multiply-adds are counted as two FLOPs because, in many recent models, convolutions are bias-free and it makes sense to count multiply and add as separate FLOPs.
\subsection{Inference time}
\label{sec:inference-time}
We report inference time per image for each DNN model for both the NVIDIA Titan X Pascal GPU and the NVIDIA Jetson TX1. We measure inference time in terms of milliseconds and by considering the same batch sizes described in Section~\ref{sec:memory}. For statistical validation the reported time corresponds to the average over 10 runs.
\begin{figure*}[!t]
\centering
\resizebox{\textwidth}{!}{
    \begin{tabular}{cc}
    \includegraphics[height=8truecm]{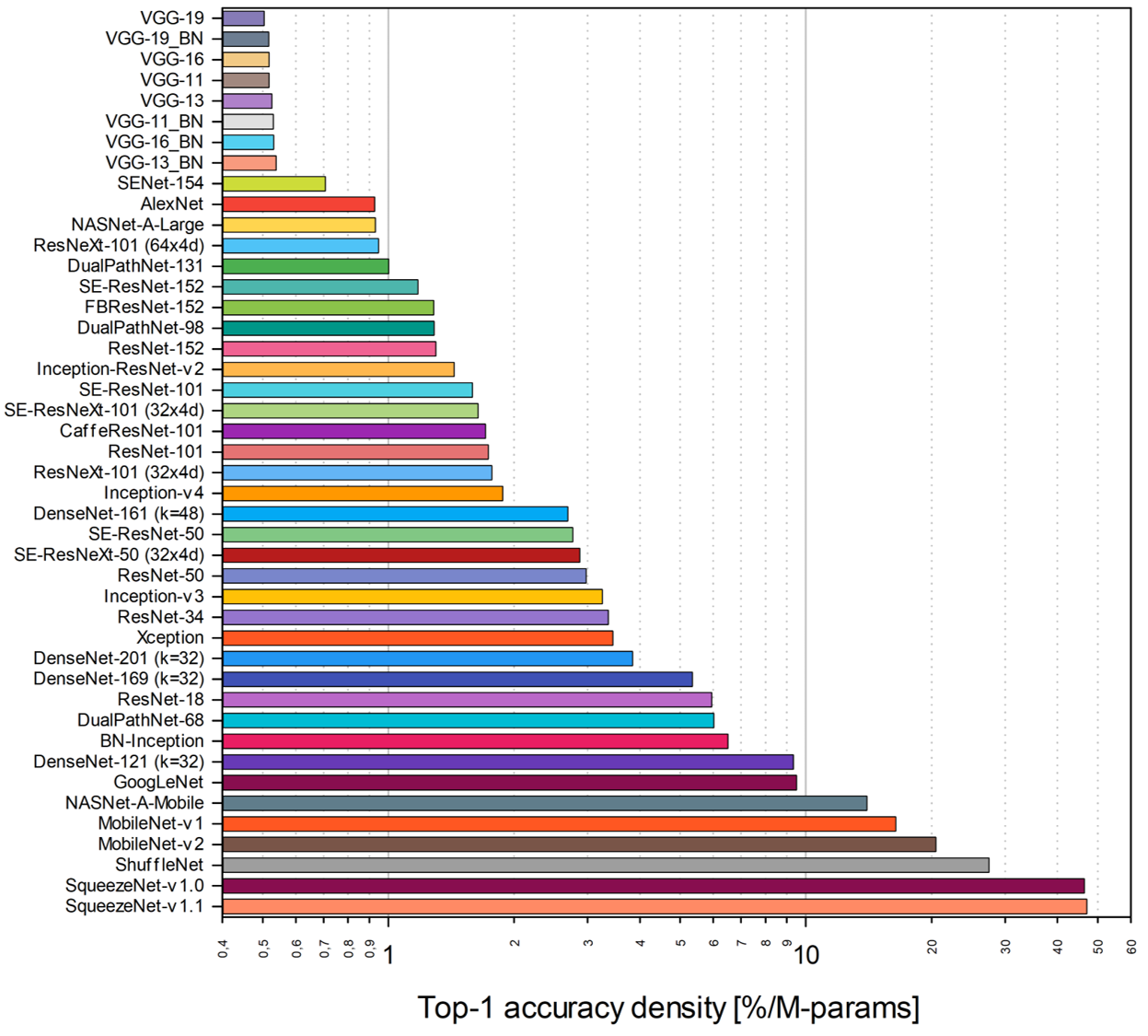} &
    \includegraphics[height=8truecm]{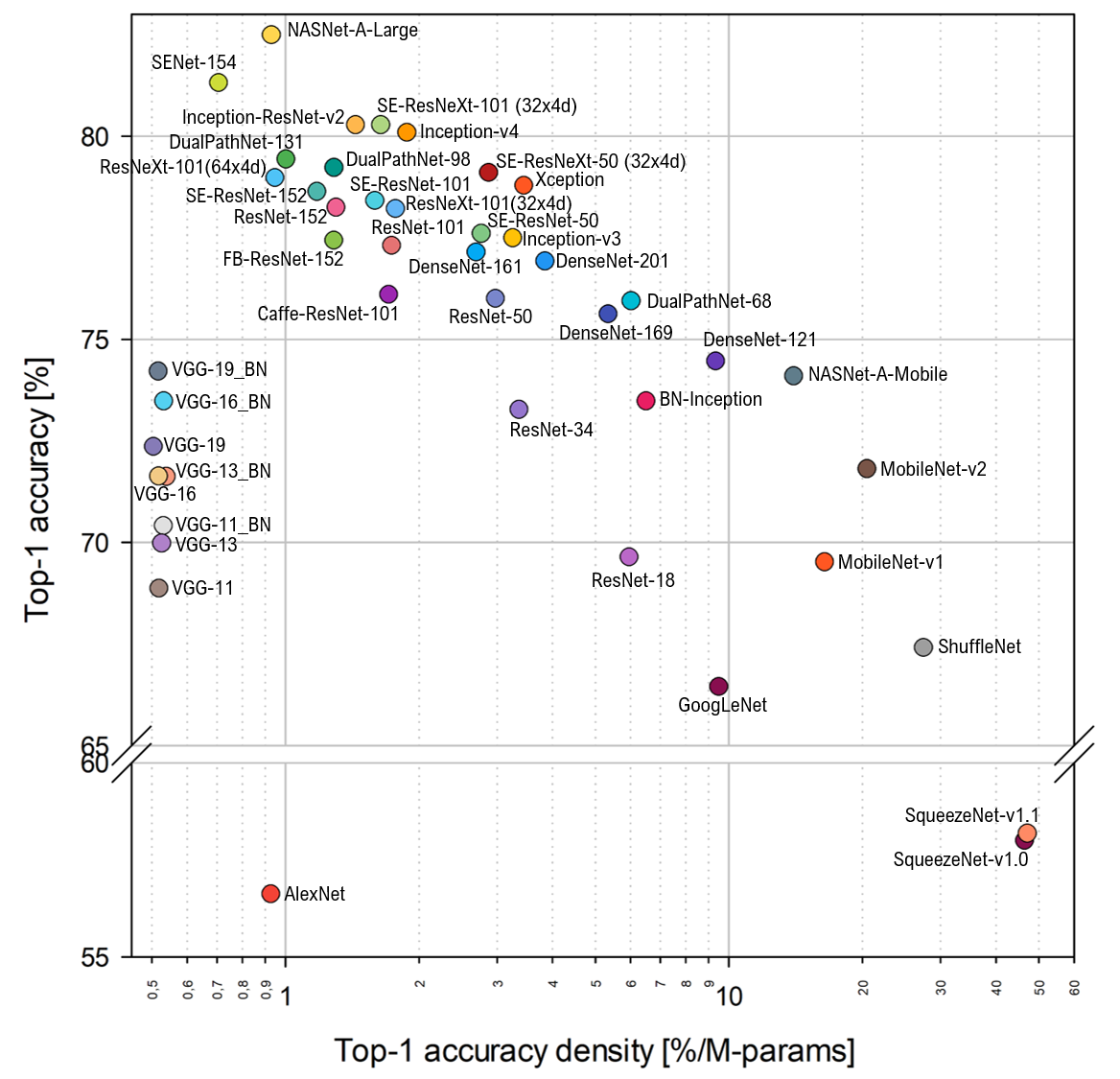} \\
    (a)&(b)\\
    \end{tabular}}
    \caption{Top-1 accuracy density (a) and Top-1 accuracy \emph{vs.} Top-1 accuracy density (b). The accuracy density measures how efficiently each model uses its parameters.}
    \label{fig:accuracydensity}
\end{figure*}

\section{Results}
\label{sec:results}
\subsection{Accuracy-rate vs Computational  Complexity vs Model Complexity}
The ball charts reported in Figures \ref{fig:accuracyvsflops} (a) and (b) show Top-1 and Top-5 accuracy on the ImageNet-1k validation set with respect to the computational complexity of the considered architectures for a 
single forward pass measured for both the workstation and the embedded board, The ball size corresponds to the model complexity. From the plots it can be seen that the DNN model reaching the highest Top-1 and Top-5 accuracy is the NASNet-A-Large that is also the one having the highest computational complexity. 
Among the models having the lowest computational complexity instead (i.e. lower than 5 G-FLOPs), SE-ResNeXt-50 (32x4d) is the one reaching the highest Top-1 and Top-5 accuracy showing at the same time a low level of model complexity, with approximately 2.76 M-params. Overall, it seems that there is no relationship between computational complexity and recognition accuracy, for instance SENet-154 needs about $3\times$ the number of operations that are needed by  SE-ResNeXt-101(32x4d) while having almost the same accuracy. Moreover, it seems that there is no relationship also between model complexity and recognition accuracy: for instance  VGG-13 has a much higher level of model complexity (size of the ball) than ResNet-18 while having almost the same accuracy. 
\subsection{Accuracy-rate vs Learning Power}
It is known that DNNs are inefficient in the use of their full learning power (measured
as the number of parameters with respect to the degrees of freedom). 
Although many papers exist that exploit this feature to produce compressed DNN models with the same accuracy of the original models \cite{han2015deep} we want here to measure how efficiently each model uses its parameters. We follow \cite{canziani2016analysis} and measure it as Top-1 accuracy density, i.e. Top-1 accuracy divided by the number of parameters. The higher is this value and the higher is the efficiency. The plot is reported in Figure~\ref{fig:accuracydensity}(a), where it can be seen that the models that use their parameters most efficiently are the SqueezeNets, ShuffleNet, the MobileNets and NASNet-A-Mobile. To focus to the density information, we plot the Top-1 accuracy with respect to the Top-1 accuracy density (see Figure~\ref{fig:accuracydensity}(b)), that permits to find more easily the desired trade-off. In this way it is possible to easily see that among the most efficient models, NASNet-A-Mobile and MobileNet-v2 are the two providing a much higher Top-1 accuracy. Among the models having the highest Top-1 accuracy (i.e. higher than 80\%) we can observe how the models using their parameters more efficiently are Inception-v4 and SE-ResNeXt-101 (32x4d).

\subsection{Inference time}

Average per image inference time over 10 runs for all the DNN models considered are reported in Tables \ref{table:inferencetime}(a) and (b) for batch size equal to 1, 2, 4, 8, 16, 32, and 64 on both the Titan Xp and the Jetson. Inference time is measured in milliseconds and the entries in Tables \ref{table:inferencetime}(a) and (b) are color coded to easily convert them in frames per second (FPS). From the table it is possible to see that all the DNN models considered are able to achieve super real-time performances on the  Titan Xp with the only exception of SENet-154, when a batch size of 1 is considered. On the Jetson instead, only a few models are able to achieve super real-time performances when a batch size of 1 is considered, namely: the SqueezeNets, the MobileNets, ResNet-18, GoogLeNet, and AlexNet. Missing measurements are due to the lack of enough system memory required to process the larger batches.

\subsection{Accuracy-rate vs Inference time}

\begin{figure*}[!tb]
\centering
\resizebox{\textwidth}{!}{
    \begin{tabular}{cc}
    \includegraphics[height=8truecm]{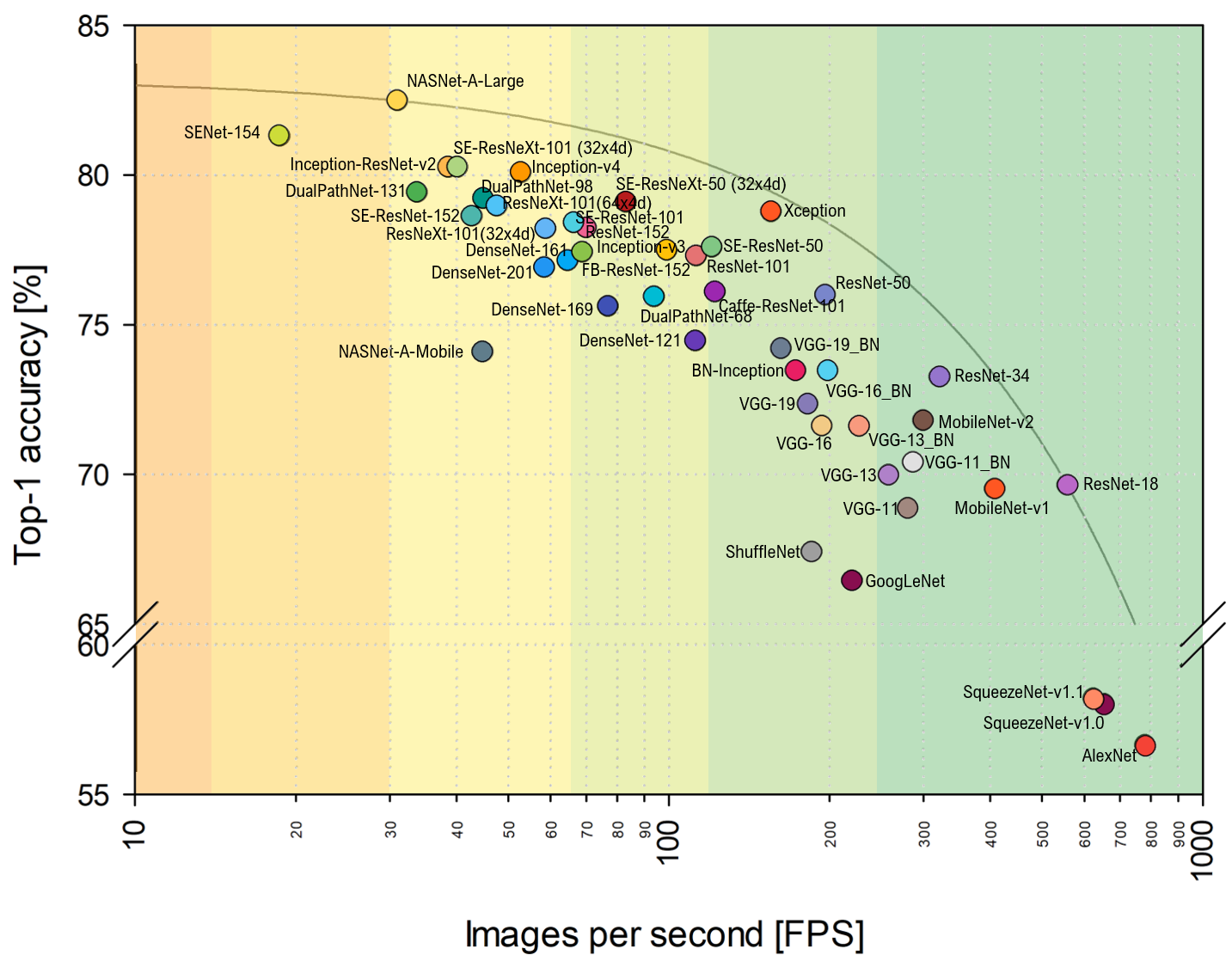} &
    \includegraphics[height=8truecm]{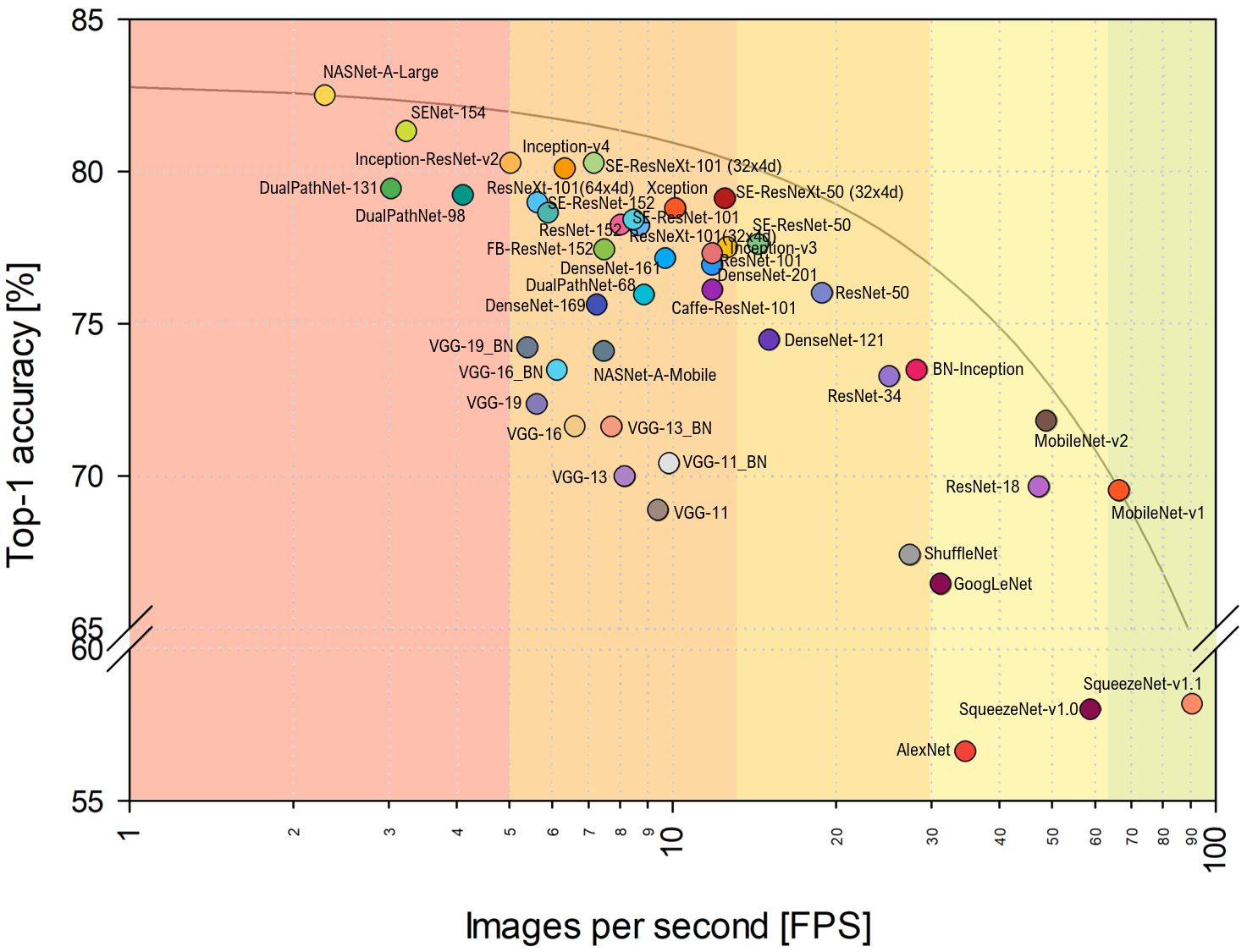} \\
    (a)&(b)\\
    \end{tabular}}
    \caption{Top-1 accuracy \textit{vs.} number of images processed per second (with batch size 1) using the  Titan Xp (a) and Jetson TX1 (b).}
    \label{fig:accuracyvsfps}
\end{figure*}

In Figure~\ref{fig:accuracyvsfps}(a) and (b) we report the plots of the top-1 accuracy with respect to the number of images processed per second (i.e. the number of inferences per second) with a batch size of 1 on both the Titan Xp and the Jetson TX1. On each plot the linear upper bound is also reported; the two have almost the same intercept ($\approx83.3$ for Titan Xp and $\approx83.0$ for the Jetson TX1), but the first has a slope that is almost $8.3\times$ smaller than the second one ($-0.0244$ \textit{vs.} $-0.2025$); these bounds show that the Titan Xp guarantees a lower decay of the maximum accuracy achievable when a larger throughput is needed. Note that this bound appears a curve instead of line in the plots because of the logarithmic scale of the images per second axis. From the Titan Xp plot it is possible to see that if one targets a throughput of more than 250 FPS, the model giving the highest accuracy is ResNet-34, with 73.27\% Top-1 accuracy; with a target of more than 125 FPS the model giving the highest accuracy is Xception, with 78,79\% Top-1 accuracy; with a target of more than 62.5 FPS the model giving the highest accuracy is SE-ResNeXt-50 (32x4d), with 79,11\% Top-1 accuracy; with a target of more than 30 FPS the model giving the highest accuracy is NASNet-A-Large, with 82,50\% Top-1 accuracy. This analysis shows how even the most accurate model in the state of the art, i.e. NASNet-A-Large, is able to provide super real-time performance (30.96 FPS) on the Titan Xp. 
Considering the Jetson TX1 plot it is possible to see that if one targets super real-time performance, the model giving the highest accuracy is MobileNet-v2, with a Top-1 accuracy of 71.81\%; if one targets a Top-1 accuracy larger than 75\%, the maximum throughput is achieved by ResNet-50 (18,83 FPS); targeting a Top-1 accuracy larger than 80\%, the maximum throughput is achieved by SE-ResNeXt-101 (32x4d) (7,16 FPS); targeting the highest Top-1 accuracy in the state of the art the throughput achievable is 2,29 FPS.
\begin{table*}[ht]
\caption{Inference time \emph{vs.} batch size. Inference time per image is estimated across different batch sizes for the Titan Xp (left), and Jetson TX1 (right). Missing data are due to the lack of enough system memory required to process the larger batches.}
\label{tab:inference}
\begin{center}
\begin{tabular}{cc}    \resizebox{\columnwidth}{0.26\textheight}{%
\begin{tabular}{rcccccccc}
DNN & 1&2&4&8&16&32&64\\
\hline\\
AlexNet &\zz{1.28} &\zz{0.70} &\zz{0.48} &\zz{0.27} &\zz{0.18} &\zz{0.14} &\zz{0.15} \\
BN-Inception &\zz{5.79} &\zz{3.00} &\zz{1.64} &\zz{1.10} &\zz{0.87} &\zz{0.77} &\zz{0.71} \\
CaffeResNet-101 &\zz{8.20} &\zz{4.82} &\zz{3.32} &\zz{2.54} &\zz{2.27} &\zz{2.16} &\zz{2.08} \\
DenseNet-121 (k=32) &\zz{8.93} &\zz{4.41} &\zz{2.64} &\zz{1.96} &\zz{1.64} &\zz{1.44} &\zz{1.39} \\
DenseNet-169 (k=32) &\zz{13.03} &\zz{6.72} &\zz{3.97} &\zz{2.73} &\zz{2.14} &\zz{1.87} &\zz{1.75} \\
DenseNet-201 (k=32) &\zz{17.15} &\zz{9.25} &\zz{5.36} &\zz{3.66} &\zz{2.84} &\zz{2.41} &\zz{2.27} \\
DenseNet-161 (k=48) &\zz{15.50} &\zz{9.10} &\zz{5.89} &\zz{4.45} &\zz{3.66} &\zz{3.43} &\zz{3.24} \\
DPN-68 &\zz{10.68} &\zz{5.36} &\zz{3.24} &\zz{2.47} &\zz{1.80} &\zz{1.59} &\zz{1.52} \\
DPN-98 &\zz{22.31} &\zz{13.84} &\zz{8.97} &\zz{6.77} &\zz{5.59} &\zz{4.96} &\zz{4.72} \\
DPN-131 &\zz{29.70} &\zz{18.29} &\zz{11.96} &\zz{9.12} &\zz{7.57} &\zz{6.72} &\zz{6.37} \\
FBResNet-152 &\zz{14.55} &\zz{7.79} &\zz{5.15} &\zz{4.31} &\zz{3.96} &\zz{3.76} &\zz{3.65} \\
GoogLeNet &\zz{4.54} &\zz{2.44} &\zz{1.65} &\zz{1.06} &\zz{0.86} &\zz{0.76} &\zz{0.72} \\
Inception-ResNet-v2 &\zz{25.94} &\zz{14.36} &\zz{8.82} &\zz{6.43} &\zz{5.19} &\zz{4.88} &\zz{4.59} \\
Inception-v3 &\zz{10.10} &\zz{5.70} &\zz{3.65} &\zz{2.54} &\zz{2.05} &\zz{1.89} &\zz{1.80} \\
Inception-v4 &\zz{18.96} &\zz{10.61} &\zz{6.53} &\zz{4.85} &\zz{4.10} &\zz{3.77} &\zz{3.61} \\
MobileNet-v1 &\zz{2.45} &\zz{0.89} &\zz{0.68} &\zz{0.60} &\zz{0.55} &\zz{0.53} &\zz{0.53} \\
MobileNet-v2 &\zz{3.34} &\zz{1.63} &\zz{0.95} &\zz{0.78} &\zz{0.72} &\zz{0.63} &\zz{0.61} \\
NASNet-A-Large &\zz{32.30} &\zz{23.00} &\zz{19.75} &\zz{18.49} &\zz{18.11} &\zz{17.73} &\zz{17.77} \\
NASNet-A-Mobile &\zz{22.36} &\zz{11.44} &\zz{5.60} &\zz{2.81} &\zz{1.61} &\zz{1.75} &\zz{1.51} \\
ResNet-101 &\zz{8.90} &\zz{5.16} &\zz{3.32} &\zz{2.69} &\zz{2.42} &\zz{2.29} &\zz{2.21} \\
ResNet-152 &\zz{14.31} &\zz{7.36} &\zz{4.68} &\zz{3.83} &\zz{3.50} &\zz{3.30} &\zz{3.17} \\
ResNet-18 &\zz{1.79} &\zz{1.01} &\zz{0.70} &\zz{0.56} &\zz{0.51} &\zz{0.41} &\zz{0.38} \\
ResNet-34 &\zz{3.11} &\zz{1.80} &\zz{1.20} &\zz{0.96} &\zz{0.82} &\zz{0.71} &\zz{0.67} \\
ResNet-50 &\zz{5.10} &\zz{2.87} &\zz{1.99} &\zz{1.65} &\zz{1.49} &\zz{1.37} &\zz{1.34} \\
ResNeXt-101 (32x4d) &\zz{17.05} &\zz{9.02} &\zz{6.27} &\zz{4.62} &\zz{3.71} &\zz{3.25} &\zz{3.11} \\
ResNeXt-101 (64x4d) &\zz{21.05} &\zz{15.54} &\zz{10.39} &\zz{7.80} &\zz{6.39} &\zz{5.62} &\zz{5.29} \\
SE-ResNet-101 &\zz{15.10} &\zz{9.26} &\zz{6.17} &\zz{4.72} &\zz{4.03} &\zz{3.62} &\zz{3.42} \\
SE-ResNet-152 &\zz{23.43} &\zz{13.08} &\zz{8.74} &\zz{6.55} &\zz{5.51} &\zz{5.06} &\zz{4.85} \\
SE-ResNet-50 &\zz{8.32} &\zz{5.16} &\zz{3.36} &\zz{2.62} &\zz{2.22} &\zz{2.01} &\zz{2.06} \\
SE-ResNeXt-101 (32x4d) &\zz{24.96} &\zz{13.86} &\zz{9.16} &\zz{6.55} &\zz{5.29} &\zz{4.53} &\zz{4.29} \\
SE-ResNeXt-50 (32x4d) &\zz{12.06} &\zz{7.41} &\zz{5.12} &\zz{3.64} &\zz{2.97} &\zz{3.01} &\zz{2.56} \\
SENet-154 &\zz{53.80} &\zz{30.30} &\zz{19.32} &\zz{13.27} &\zz{10.45} &\zz{9.41} &\zz{8.91} \\
ShuffleNet &\zz{5.40} &\zz{2.67} &\zz{1.37} &\zz{0.82} &\zz{0.66} &\zz{0.59} &\zz{0.56} \\
SqueezeNet-v1.0 &\zz{1.53} &\zz{0.84} &\zz{0.66} &\zz{0.59} &\zz{0.54} &\zz{0.52} &\zz{0.53} \\
SqueezeNet-v1.1 &\zz{1.60} &\zz{0.77} &\zz{0.44} &\zz{0.37} &\zz{0.32} &\zz{0.31} &\zz{0.30} \\
VGG-11 &\zz{3.57} &\zz{4.40} &\zz{2.89} &\zz{1.56} &\zz{1.19} &\zz{1.10} &\zz{1.13} \\
VGG-11\_BN &\zz{3.49} &\zz{4.60} &\zz{2.99} &\zz{1.71} &\zz{1.33} &\zz{1.24} &\zz{1.27} \\
VGG-13 &\zz{3.88} &\zz{5.03} &\zz{3.44} &\zz{2.25} &\zz{1.83} &\zz{1.75} &\zz{1.79} \\
VGG-13\_BN &\zz{4.40} &\zz{5.37} &\zz{3.71} &\zz{2.42} &\zz{2.05} &\zz{1.97} &\zz{2.00} \\
VGG-16 &\zz{5.17} &\zz{5.91} &\zz{4.01} &\zz{2.84} &\zz{2.20} &\zz{2.12} &\zz{2.15} \\
VGG-16\_BN &\zz{5.04} &\zz{5.95} &\zz{4.27} &\zz{3.06} &\zz{2.45} &\zz{2.36} &\zz{2.41} \\
VGG-19 &\zz{5.50} &\zz{6.26} &\zz{4.71} &\zz{3.29} &\zz{2.59} &\zz{2.52} &\zz{2.50} \\
VGG-19\_BN &\zz{6.17} &\zz{6.67} &\zz{4.86} &\zz{3.56} &\zz{2.88} &\zz{2.74} &\zz{2.76} \\
Xception &\zz{6.44} &\zz{5.35} &\zz{4.90} &\zz{4.47} &\zz{4.41} &\zz{4.41} &\zz{4.36} \\
 && & & & & & \\
\hline\\
                \end{tabular}
                }
         & 
         \resizebox{\columnwidth}{0.26\textheight}{%
    \begin{tabular}{rcccccccc}
DNN & 1&2&4&8&16&32&64\\
\hline\\
AlexNet &\zz{28.88} &\zz{13.00} &\zz{8.58} &\zz{6.56} &\zz{5.39} &\zz{4.77} &  \\
BN-Inception &\zz{35.52} &\zz{26.48} &\zz{25.10} &\zz{23.89} &\zz{21.21} &\zz{20.47} &  \\
CaffeResNet-101 &\zz{84.47} &\zz{91.37} &\zz{70.33} &\zz{63.53} &\zz{56.38} &\zz{53.73} &  \\
DenseNet-121 (k=32) &\zz{66.43} &\zz{50.87} &\zz{50.20} &\zz{43.89} &\zz{40.41} &\zz{38.22} &  \\
DenseNet-169 (k=32) &\zz{137.96} &\zz{130.27} &\zz{110.82} &\zz{100.56} &\zz{92.97} &\zz{88.94} &  \\
DenseNet-201 (k=32) &\zz{84.57} &\zz{61.71} &\zz{62.62} &\zz{53.73} &\zz{49.28} &\zz{46.26} &  \\
DenseNet-161 (k=48) &\zz{103.20} &\zz{76.11} &\zz{77.10} &\zz{68.32} &\zz{62.73} &\zz{59.14} &  \\
DPN-68 &\zz{113.08} &\zz{52.73} &\zz{42.85} &\zz{43.32} &\zz{38.18} &\zz{36.40} &\zz{36.22} \\
DPN-98 &\zz{243.51} &\zz{148.51} &\zz{135.30} &\zz{125.92} &\zz{123.34} &\zz{118.68} &\zz{117.27} \\
DPN-131 &\zz{330.15} &\zz{204.69} &\zz{184.89} &\zz{172.25} &\zz{165.59} &\zz{162.67} &\zz{160.66} \\
FBResNet-152 &\zz{133.68} &\zz{147.75} &\zz{113.48} &\zz{105.78} &\zz{94.26} &\zz{97.47} &  \\
GoogLeNet &\zz{32.11} &\zz{27.19} &\zz{23.29} &\zz{21.66} &\zz{19.77} &\zz{19.96} &  \\
Inception-ResNet-v2 &\zz{198.95} &\zz{141.29} &\zz{127.97} &\zz{130.25} &\zz{117.99} &\zz{116.47} &  \\
Inception-v3 &\zz{79.39} &\zz{59.04} &\zz{56.46} &\zz{51.79} &\zz{47.60} &\zz{46.85} &  \\
Inception-v4 &\zz{158.00} &\zz{120.23} &\zz{106.77} &\zz{102.21} &\zz{95.51} &\zz{95.40} &  \\
MobileNet-v1 &\zz{15.06} &\zz{11.94} &\zz{11.34} &\zz{11.03} &\zz{10.82} &\zz{10.58} &\zz{10.55} \\
MobileNet-v2 &\zz{20.51} &\zz{14.58} &\zz{13.67} &\zz{13.56} &\zz{13.18} &\zz{13.10} &\zz{12.72} \\
NASNet-A-Large &\zz{437.20} &\zz{399.99} &\zz{385.75} &\zz{383.55} &\zz{389.67} &  &  \\
NASNet-A-Mobile &\zz{133.87} &\zz{62.91} &\zz{33.72} &\zz{30.62} &\zz{29.72} &\zz{28.92} &\zz{28.55} \\
ResNet-101 &\zz{84.52} &\zz{77.90} &\zz{71.23} &\zz{67.14} &\zz{58.11} &  &  \\
ResNet-152 &\zz{124.67} &\zz{113.65} &\zz{101.41} &\zz{96.76} &\zz{82.35} &  &  \\
ResNet-18 &\zz{21.16} &\zz{15.30} &\zz{14.35} &\zz{13.82} &\zz{11.99} &\zz{10.73} &\zz{12.45} \\
ResNet-34 &\zz{39.88} &\zz{28.82} &\zz{27.51} &\zz{24.97} &\zz{20.41} &\zz{18.48} &\zz{17.97} \\
ResNet-50 &\zz{53.09} &\zz{44.84} &\zz{41.20} &\zz{38.79} &\zz{35.72} &  &  \\
ResNeXt-101 (32x4d) &\zz{115.37} &\zz{90.93} &\zz{84.64} &\zz{79.66} &\zz{77.63} &  &  \\
ResNeXt-101 (64x4d) &\zz{177.40} &\zz{155.77} &\zz{144.82} &\zz{137.43} &\zz{134.07} &  &  \\
SE-ResNet-101 &\zz{118.13} &\zz{105.11} &\zz{96.71} &\zz{91.68} &\zz{80.99} &  &  \\
SE-ResNet-152 &\zz{169.73} &\zz{155.08} &\zz{139.72} &\zz{133.59} &\zz{116.97} &  &  \\
SE-ResNet-50 &\zz{69.65} &\zz{61.37} &\zz{55.33} &\zz{51.87} &\zz{47.80} &  &  \\
SE-ResNeXt-101 (32x4d) &\zz{139.62} &\zz{122.01} &\zz{112.05} &\zz{105.34} &\zz{102.39} &  &  \\
SE-ResNeXt-50 (32x4d) &\zz{80.08} &\zz{69.86} &\zz{67.20} &\zz{62.66} &\zz{61.19} &  &  \\
SENet-154 &\zz{309.48} &\zz{240.80} &\zz{221.84} &\zz{211.00} &\zz{207.06} &\zz{201.49} &\zz{201.66} \\
ShuffleNet &\zz{36.58} &\zz{22.61} &\zz{13.80} &\zz{13.36} &\zz{12.91} &\zz{12.66} &\zz{12.50} \\
SqueezeNet-v1.0 &\zz{17.00} &\zz{16.47} &\zz{15.03} &\zz{13.97} &\zz{13.25} &\zz{12.89} &\zz{12.70} \\
SqueezeNet-v1.1 &\zz{11.05} &\zz{9.88} &\zz{8.80} &\zz{7.90} &\zz{7.38} &\zz{7.20} &\zz{7.04} \\
VGG-11 &\zz{106.44} &\zz{125.84} &\zz{85.84} &\zz{60.10} &\zz{32.56} &\zz{30.51} &\zz{32.27} \\
VGG-11\_BN &\zz{101.58} &\zz{122.82} &\zz{86.26} &\zz{54.50} &\zz{47.81} &\zz{47.31} &\zz{41.26} \\
VGG-13 &\zz{122.59} &\zz{148.80} &\zz{108.28} &\zz{75.99} &\zz{70.57} &\zz{64.88} &\zz{62.79} \\
VGG-13\_BN &\zz{129.69} &\zz{153.68} &\zz{113.90} &\zz{81.19} &\zz{76.39} &\zz{70.59} &\zz{67.38} \\
VGG-16 &\zz{151.52} &\zz{169.92} &\zz{129.89} &\zz{96.81} &\zz{91.72} &  &  \\
VGG-16\_BN &\zz{163.37} &\zz{176.35} &\zz{136.85} &\zz{103.45} &\zz{98.11} &  &  \\
VGG-19 &\zz{178.04} &\zz{192.86} &\zz{152.28} &\zz{117.92} &\zz{112.39} &  &  \\
VGG-19\_BN &\zz{185.18} &\zz{198.66} &\zz{159.20} &\zz{124.88} &\zz{119.15} &  &  \\
Xception &\zz{98.96} &\zz{93.40} &\zz{90.49} &\zz{87.65} &\zz{86.89} &  &  \\
 && & & & & & \\
\hline\\
            \end{tabular}
            } \\
            &\\
            (a)&(b)\\
    \multicolumn{2}{c}{
    %\resizebox{.7\columnwidth}{14pt}{%
    \scriptsize
        \begin{tabular}{p{13pt}p{13pt}p{13pt}p{13pt}p{13pt}p{13pt}p{13pt}p{13pt}p{13pt}}
        &&&&&&&&\\
        FPS&>1000&>250&>125&>62.5&>30&>15&>5&<=5\\
        &\cellcolor{teal}&\cellcolor{green}&\cellcolor{greenlight}&\cellcolor{lime}&\cellcolor{yellow1}&\cellcolor{amber}&\cellcolor{orange}&\cellcolor{deeporange}\\
        ms&<1&<4&<8&<16&<33&<66&<200&>=200\\
        \end{tabular}
    }%}
    \end{tabular}
\label{table:inferencetime}
\end{center}
\end{table*}

\subsection{Memory usage}

In Table \ref{tab:memory} we analyze the memory consumption for all the DNN models considered for different batch sizes on the Titan Xp. From the memory footprints reported it can be seen that when a batch size of 1 is considered, most models require less than 1GB of memory, with the exception of NASNet-A-Large, the SE-ResNets, the SE-ResNeXTs, SENet-154, the VGGs and Xception. However none of them requires more than 1.5GB for a batch size of 1.

\subsection{Memory usage vs Model Complexity}

\begin{figure}[!t]
\centering
\resizebox{\columnwidth}{!}{
    \includegraphics[height=8truecm]{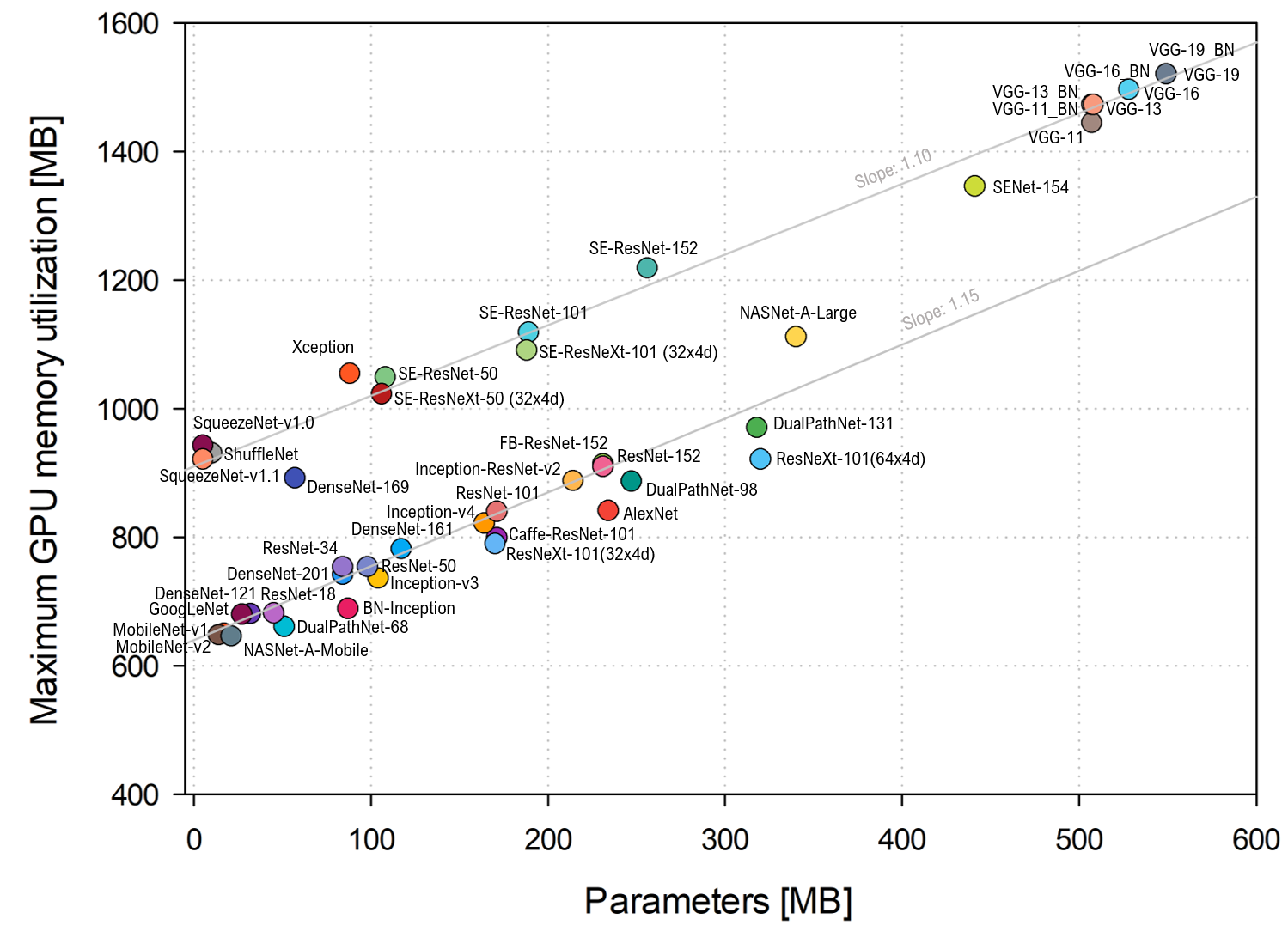} }
    \caption{Plot of the initial static allocation of the model parameters (i.e. the model complexity) and the total memory utilization with batch size 1 on the Titan Xp.}
    \label{fig:linee}
\end{figure}

In Figure \ref{fig:linee} we analyze the relationship between the initial static allocation of the model parameters (i.e. the model complexity) and the total memory utilization for a batch size of 1 on the Titan Xp. 
We can see that the relationship is linear, and follows two different lines with approximately the same slope (i.e. 1.10 and 1.15) and different intercepts (i.e. 910 and 639 respectively). This means that the model complexity can be used to reliably estimate the total memory utilization. From the plots we can observe that families of models belong to the same line, as for example the VGGs, the SE-ResNets ans SqueezeNets lie on the line with higher intercept, while the ResNets, DualPathNets, DenseNets, Inception nets and MobileNets line on the line with the lower intercept. In particular we can observe how models having the smallest complexity (i.e. SqueezeNet-v1.0 and SqueezeNet-v1.1 both with 5MB) have a 943MB and 921MB memory footprint, while models having slightly higher complexity (i.e. MobileNet-v1 and MobileNet-v2 with respectively 17MB and 14MB) have a much smaller memory footprint, equal to 650MB and 648MB respectively. 
\subsection{Best DNN at given constraints}
Table \ref{tab:constraints} shows the best DNN architectures in terms of recognition accuracy when specific hardware resources are given as computational constraints. This analysis is done for both the Titan Xp and  Jetson TX1.
We define the following constraints:
\begin{itemize}
    \item[-] Memory usage: high ($\leq$1.4GB), medium ($\leq$1.0GB) and low ($\leq$0.7GB);
    \item[-] Computational time: half real-time (@15FPS), real-time (@30FPS), super real-time (@60FPS);
\end{itemize}
 A Titan Xp, with a low memory usage as constraint, achieves a recognition accuracy of at most 75.95\% by using the DPN-68 network independently of the computational time. Having more resources, for instance medium and high memory usage, Titan Xp achieves a recognition accuracy of at most 79.11\% by using the SE-ResNeXt-50 (32x4d) with a super real-time throughput. Having  no requirements in terms of memory usage, the Jetson TX1 achieves a recognition accuracy of at most 69.52\% by using the MobileNet-v1, which guarantees a super real-time throughput. To have a DNN running on a Jetson that is comparable in terms of recognition accuracy to the best DNNs running on the Titan Xp, a memory size of at least 1GB is needed. In this case the most performing is the ResNet-50, able to guarantee an half real-time throughput, with a recognition accuracy of 76.01\%.

\begin{table}[!htbp]
    \begin{center}
    \caption{Memory consumption of the different DNN models considered on the Titan Xp for different batch sizes.}
    \label{tab:memory}
\begin{tabular}{c}   
\resizebox{\columnwidth}{0.26\textheight}{%
        \begin{tabular}{rccccccc}
DNN & 1&2&4&8&16&32&64\\
\hline\\
AlexNet &\mem{0.82} &\mem{0.83} &\mem{0.83} &\mem{0.83} &\mem{0.84} &\mem{0.92} &\mem{0.99} \\
BN-Inception &\mem{0.67} &\mem{0.71} &\mem{0.81} &\mem{0.97} &\mem{1.29} &\mem{1.97} &\mem{3.24} \\
CaffeResNet-101 &\mem{0.78} &\mem{0.80} &\mem{0.80} &\mem{0.80} &\mem{0.81} &\mem{0.95} &\mem{1.06} \\
DenseNet-121 (k=32) &\mem{0.67} &\mem{0.67} &\mem{0.67} &\mem{0.69} &\mem{0.71} &\mem{0.71} &\mem{0.99} \\
DenseNet-169 (k=32) &\mem{0.87} &\mem{0.87} &\mem{0.88} &\mem{0.91} &\mem{0.93} &\mem{0.97} &\mem{1.04} \\
DenseNet-201 (k=32) &\mem{0.72} &\mem{0.72} &\mem{0.73} &\mem{0.75} &\mem{0.77} &\mem{0.80} &\mem{0.87} \\
DenseNet-161 (k=48) &\mem{0.76} &\mem{0.77} &\mem{0.77} &\mem{0.80} &\mem{0.82} &\mem{0.88} &\mem{0.96} \\
DPN-68 &\mem{0.65} &\mem{0.65} &\mem{0.66} &\mem{0.67} &\mem{0.67} &\mem{0.68} &\mem{0.71} \\
DPN-98 &\mem{0.87} &\mem{0.88} &\mem{0.90} &\mem{0.92} &\mem{0.98} &\mem{1.10} &\mem{1.29} \\
DPN-131 &\mem{0.95} &\mem{0.95} &\mem{0.96} &\mem{0.97} &\mem{1.05} &\mem{1.04} &\mem{1.28} \\
FBResNet-152 &\mem{0.89} &\mem{0.90} &\mem{0.92} &\mem{0.94} &\mem{0.97} &\mem{1.12} &\mem{1.31} \\
GoogLeNet &\mem{0.66} &\mem{0.70} &\mem{0.76} &\mem{0.87} &\mem{1.09} &\mem{1.51} &\mem{2.35} \\
Inception-ResNet-v2 &\mem{0.87} &\mem{0.88} &\mem{0.88} &\mem{0.89} &\mem{0.91} &\mem{0.95} &\mem{1.02} \\
Inception-v3 &\mem{0.72} &\mem{0.73} &\mem{0.75} &\mem{0.75} &\mem{0.77} &\mem{0.83} &\mem{0.92} \\
Inception-v4 &\mem{0.80} &\mem{0.81} &\mem{0.82} &\mem{0.84} &\mem{0.90} &\mem{0.90} &\mem{1.18} \\
MobileNet-v1 &\mem{0.63} &\mem{0.64} &\mem{0.64} &\mem{0.65} &\mem{0.67} &\mem{0.71} &\mem{0.78} \\
MobileNet-v2 &\mem{0.63} &\mem{0.63} &\mem{0.63} &\mem{0.64} &\mem{0.66} &\mem{0.70} &\mem{0.78} \\
NASNet-A-Large &\mem{1.09} &\mem{1.19} &\mem{1.38} &\mem{1.78} &\mem{2.56} &\mem{4.12} &\mem{7.26} \\
NASNet-A-Mobile &\mem{0.63} &\mem{0.65} &\mem{0.67} &\mem{0.71} &\mem{0.79} &\mem{0.93} &\mem{1.23} \\
ResNet-101 &\mem{0.82} &\mem{0.83} &\mem{0.86} &\mem{0.93} &\mem{1.08} &\mem{1.37} &\mem{1.94} \\
ResNet-152 &\mem{0.89} &\mem{0.90} &\mem{0.92} &\mem{1.00} &\mem{1.15} &\mem{1.43} &\mem{2.01} \\
ResNet-18 &\mem{0.67} &\mem{0.68} &\mem{0.68} &\mem{0.69} &\mem{0.71} &\mem{0.75} &\mem{0.89} \\
ResNet-34 &\mem{0.74} &\mem{0.74} &\mem{0.75} &\mem{0.80} &\mem{0.90} &\mem{1.09} &\mem{1.47} \\
ResNet-50 &\mem{0.74} &\mem{0.74} &\mem{0.77} &\mem{0.85} &\mem{0.99} &\mem{1.28} &\mem{1.86} \\
ResNeXt-101 (32x4d) &\mem{0.77} &\mem{0.78} &\mem{0.78} &\mem{0.79} &\mem{0.84} &\mem{0.87} &\mem{1.06} \\
ResNeXt-101 (64x4d) &\mem{0.90} &\mem{0.92} &\mem{0.91} &\mem{0.96} &\mem{1.01} &\mem{1.19} &\mem{1.38} \\
SE-ResNet-101 &\mem{1.09} &\mem{1.11} &\mem{1.10} &\mem{1.13} &\mem{1.13} &\mem{1.27} &\mem{1.36} \\
SE-ResNet-152 &\mem{1.19} &\mem{1.21} &\mem{1.25} &\mem{1.35} &\mem{1.54} &\mem{1.93} &\mem{2.69} \\
SE-ResNet-50 &\mem{1.02} &\mem{1.04} &\mem{1.08} &\mem{1.18} &\mem{1.38} &\mem{1.76} &\mem{2.53} \\
SE-ResNeXt-101 (32x4d) &\mem{1.07} &\mem{1.07} &\mem{1.08} &\mem{1.09} &\mem{1.12} &\mem{1.16} &\mem{1.25} \\
SE-ResNeXt-50 (32x4d) &\mem{1.00} &\mem{1.03} &\mem{1.08} &\mem{1.19} &\mem{1.38} &\mem{1.76} &\mem{2.53} \\
SENet-154 &\mem{1.31} &\mem{1.32} &\mem{1.33} &\mem{1.36} &\mem{1.40} &\mem{1.48} &\mem{1.65} \\
ShuffleNet &\mem{0.91} &\mem{0.91} &\mem{0.92} &\mem{0.93} &\mem{0.95} &\mem{0.99} &\mem{1.05} \\
SqueezeNet-v1.0 &\mem{0.92} &\mem{0.92} &\mem{0.92} &\mem{0.93} &\mem{0.94} &\mem{0.97} &\mem{1.02} \\
SqueezeNet-v1.1 &\mem{0.90} &\mem{0.90} &\mem{0.91} &\mem{0.92} &\mem{0.94} &\mem{0.99} &\mem{1.07} \\
VGG-11 &\mem{1.41} &\mem{1.43} &\mem{1.43} &\mem{1.43} &\mem{1.53} &\mem{1.55} &\mem{1.81} \\
VGG-11\_BN &\mem{1.44} &\mem{1.49} &\mem{1.59} &\mem{1.78} &\mem{2.39} &\mem{3.59} &\mem{5.99} \\
VGG-13 &\mem{1.44} &\mem{1.43} &\mem{1.51} &\mem{1.60} &\mem{2.02} &\mem{2.41} &\mem{3.99} \\
VGG-13\_BN &\mem{1.44} &\mem{1.49} &\mem{1.59} &\mem{1.78} &\mem{2.39} &\mem{3.59} &\mem{5.99} \\
VGG-16 &\mem{1.46} &\mem{1.51} &\mem{1.61} &\mem{1.80} &\mem{2.41} &\mem{3.61} &\mem{6.02} \\
VGG-16\_BN &\mem{1.46} &\mem{1.51} &\mem{1.61} &\mem{1.80} &\mem{2.41} &\mem{3.61} &\mem{6.02} \\
VGG-19 &\mem{1.49} &\mem{1.54} &\mem{1.63} &\mem{1.83} &\mem{2.43} &\mem{3.64} &\mem{6.04} \\
VGG-19\_BN &\mem{1.49} &\mem{1.54} &\mem{1.63} &\mem{1.83} &\mem{2.43} &\mem{3.64} &\mem{6.04} \\
Xception &\mem{1.03} &\mem{1.05} &\mem{1.06} &\mem{1.08} &\mem{1.16} &\mem{1.24} &\mem{1.53} \\
 && & & & & & \\
\hline\\
        \end{tabular}
    }\\
    % \multicolumn{2}{c}{
    \resizebox{.7\columnwidth}{14pt}{%
    %\scriptsize
        \begin{tabular}{p{13pt}p{13pt}p{13pt}p{13pt}p{13pt}p{13pt}p{13pt}p{13pt}p{13pt}}
        &&&&&&&&\\
        GB&<1&<2&<3&<4&<5&<6&<7&>=7\\
        &\cellcolor{teal}&\cellcolor{green}&\cellcolor{greenlight}&\cellcolor{lime}&\cellcolor{yellow1}&\cellcolor{amber}&\cellcolor{orange}&\cellcolor{deeporange}\\
       % ms&<1&&&&<33&&&>200\\
        \end{tabular}
    }\\
    %
    %}
            \end{tabular}

    \end{center}
\end{table}

\begin{table*}[!htbp]
\caption{Top 5 models (sorted in decreasing Top-1 accuracy) satisfying memory consumption  ($\leq$0.7GB, $\leq$1.0GB, $\leq$1.4GB) and inference speed ($\geq$15FPS, $\geq$30FPS, $\geq$60FPS) constraints on the Titan Xp (a) and Jetson TX1 (b).}
\label{tab:constraints}
\centering
\resizebox{\textwidth}{!}{
\begin{tabular}{|p{2.86cm}p{0.48cm}|p{2.86cm}p{0.48cm}|p{2.80cm}p{0.48cm}||p{2.7cm}p{0.48cm}|p{2.3cm}p{0.48cm}|p{2.3cm}p{0.48cm}|}
\hline
\multicolumn{6}{|c||}{Titan Xp} & \multicolumn{6}{c|}{Jetson} \\
\hline \hline
\cellcolor{mygray}$\leq$ 0.7GB, @15FPS & \cellcolor{mygray}Acc.& \cellcolor{mygray}$\leq$ 0.7GB, @30FPS & \cellcolor{mygray}Acc.& \cellcolor{mygray}$\leq$ 0.7GB, @60FPS & \cellcolor{mygray}Acc. &
\cellcolor{mygray}$\leq$ 0.7GB, @15FPS & \cellcolor{mygray}Acc. &\cellcolor{mygray}$\leq$ 0.7GB, @30FPS & \cellcolor{mygray}Acc.& \cellcolor{mygray}$\leq$ 0.7GB, @60FPS & \cellcolor{mygray}Acc.\\
DPN-68 & 75.95 & DPN-68 & 75.95 &DPN-68 & 75.95 & DenseNet-121 (k=32) & 74.47 & MobileNet-v2 & 71.81 & MobileNet-v1 & 69.52 \\ 
	DenseNet-121 (k=32) & 74.47 & DenseNet-121 (k=32) & 74.47 & DenseNet-121 (k=32) & 74.47 & BN-Inception & 73.48 & ResNet-18 & 69.64 &  & \\ 
	NASNet-A-Mobile & 74.10 & NASNet-A-Mobile & 74.10 & BN-Inception & 73.48 & MobileNet-v2 & 71.81 & MobileNet-v1 & 69.52 & & \\ 
	BN-Inception & 73.48 & BN-Inception & 73.48 & MobileNet-v2 & 71.81 & ResNet-18 & 69.64 & GoogLeNet & 66.45 & & \\ 
	MobileNet-v2 & 71.81 & MobileNet-v2 & 71.81 & ResNet-18 & 69.64 & MobileNet-v1 & 69.52 &  &  & & \\ \hline
\cellcolor{mygray}$\leq$ 1.0GB, @15FPS & \cellcolor{mygray}Acc. & \cellcolor{mygray}$\leq$ 1.0GB, @30FPS & \cellcolor{mygray}Acc. & \cellcolor{mygray}$\leq$ 1.0GB, @60FPS & \cellcolor{mygray}Acc. &
\cellcolor{mygray}$\leq$ 1.0GB, @15FPS & \cellcolor{mygray}Acc. & \cellcolor{mygray}$\leq$ 1.0GB, @30FPS & \cellcolor{mygray}Acc. & \cellcolor{mygray}$\leq$ 1.0GB, @60FPS  & \cellcolor{mygray}Acc.\\
Inception-ResNet-v2 & 80.28 & Inception-ResNet-v2 & 80.28 & SE-ResNeXt-50(32x4d) & 79.11 & ResNet-50 & 76.01 & MobileNet-v2 & 71.81 & MobileNet-v1 & 69.52 \\ 
	Inception-v4 & 80.10 & Inception-v4 & 80.10 & ResNet-152 & 78.25 & DenseNet-121 (k=32) & 74.47 & ResNet-18 & 69.64 & SqueezeNet-v1.1 & 58.18 \\ 
	DPN-131 & 79.44 & DPN-131 & 79.44 & Inception-v3 & 77.50 & BN-Inception & 73.48 & MobileNet-v1 & 69.52 & & \\ 
	DPN-98 & 79.23 & DPN-98 & 79.23 & FBResNet-152 & 77.44 & ResNet-34 & 73.27 & GoogLeNet & 66.45 &  & \\ 
	SE-ResNeXt-50(32x4d) & 79.11 & SE-ResNeXt-50(32x4d) & 79.11 & ResNet-101 & 77.31 & MobileNet-v2 & 71.81 & SqueezeNet-v1.1 & 58.18 & & \\ 
\hline
\cellcolor{mygray}$\leq$ 1.4GB, @15FPS & \cellcolor{mygray}Acc.& \cellcolor{mygray}$\leq$ 1.4GB, @30FPS & \cellcolor{mygray}Acc.& \cellcolor{mygray}$\leq$ 1.4GB, @60FPS & \cellcolor{mygray}Acc. &
\cellcolor{mygray}$\leq$ 1.4GB, @15FPS & \cellcolor{mygray}Acc.& \cellcolor{mygray}$\leq$ 1.4GB, @30FPS & \cellcolor{mygray}Acc.& \cellcolor{mygray}$\leq$ 1.4GB, @60FPS & \cellcolor{mygray}Acc.\\
NASNet-A-Large & 82.50 & NASNet-A-Large & 82.50 & SE-ResNeXt-50(32x4d) & 79.11 & ResNet-50 & 76.01 & MobileNet-v2 & 71.81 & MobileNet-v1 & 69.52 \\ 
	SENet-154 & 81.32 & Inception-ResNet-v2 & 80.28 & Xception & 78.79 & DenseNet-121 (k=32) & 74.47 & ResNet-18 & 69.64 & SqueezeNet-v1.1 & 58.18 \\ 
	Inception-ResNet-v2 & 80.28 & SE-ResNeXt-101(32x4d) & 80.28 & SE-ResNet-101 & 78.42 & BN-Inception & 73.48 & MobileNet-v1 & 69.52 & &  \\ 
	SE-ResNeXt-101(32x4d) & 80.28 & Inception-v4 & 80.10 & ResNet-152 & 78.25 & ResNet-34 & 73.27 & GoogLeNet & 66.45 & &  \\ 
	Inception-v4 & 80.10 & DPN-131 & 79.44 & SE-ResNet-50 & 77.61 & MobileNet-v2 & 71.81 & SqueezeNet-v1.1 & 58.18 & &  \\ \hline
\end{tabular}}
\end{table*}

\section{Conclusion}
\label{sec:discussion}
The design of Deep neural networks (DNNs) with increasing complexity able to improve the performance of the ImageNet-1k competition plays a central rule in advancing the state-of-the-art also on other vision tasks. In this article we present a comparison between different DNNs in order to provide an immediate and comprehensive tool for guiding in the selection of the appropriate architecture responding to resource constraints in practical deployments and applications. Specifically, we analyze more than 40 state-of-the-art DNN architectures trained on ImageNet-1k in terms of accuracy, number of parameters, memory usage, computational complexity, and inference time.

The key findings of this paper are the following:
\begin{itemize}
    \item[-] the recognition accuracy does not increase as the number of operations increases: in fact,  there are some architectures that with a relatively low number of operations, such as the SE-ResNeXt-50 (32x4d), achieve very high accuracy (see Figures \ref{fig:accuracyvsflops}a and b). This finding is independent on the computer architecture experimented;
    \item[-] there is not a linear relationship between model complexity and accuracy (see Figures \ref{fig:accuracyvsflops}a and b);
    \item[-] not all the DNN models use their parameters with the same level of efficiency (see Figures \ref{fig:accuracydensity}a and b);
    \item[-] the desired throughput (expressed for example as the number of inferences per second) places an upper bound to the achievable accuracy (see Figures \ref{fig:accuracyvsfps}a and b);
    \item[-] model complexity can be used to reliably estimate the total memory utilization (see Figure \ref{fig:linee});
    \item[-] almost all models are capable of real-time or super real-time performance on a high-end GPU, while just a few of them can guarantee them on an embedded system (see Tables \ref{tab:inference}a and b);
%    \item[-] almost all the DNNs run at least 30 FPS operations on the Titan Xp while just some DNNs runs real-time on the Jetson TX1 (see Tables \ref{tab:inference}a and b).
    \item[-] even DNNs with a very low level model complexity have a minimum GPU memory footprint of about 0.6GB (see  Table \ref{tab:memory}).
  %  \item older DNNs (e.g. AlexNet, VGGs) run at an higher frame-rate on a recent top-of-the-line consumer GPU (i.e. Titan X) than on the Jetson TX1 (optimization of simple architectures' building blocks?)
\end{itemize}

All the DNNs considered, as well as the software used for the analysis, are available online \cite{github_celona}. We plan to add to the repository interactive plots that allow other researchers to better explore the results of this study, and more importantly to effortlessly add new entries.
\section*{Acknowledgments}
We gratefully acknowledge the support of NVIDIA Corporation with the donation of the Titan X Pascal GPU used for this research.

\bibliographystyle{IEEEtran}
\bibliography{references}

\begin{thebibliography}{10}
\providecommand{\url}[1]{#1}
\csname url@samestyle\endcsname
\providecommand{\newblock}{\relax}
\providecommand{\bibinfo}[2]{#2}
\providecommand{\BIBentrySTDinterwordspacing}{\spaceskip=0pt\relax}
\providecommand{\BIBentryALTinterwordstretchfactor}{4}
\providecommand{\BIBentryALTinterwordspacing}{\spaceskip=\fontdimen2\font plus
\BIBentryALTinterwordstretchfactor\fontdimen3\font minus
  \fontdimen4\font\relax}
\providecommand{\BIBforeignlanguage}[2]{{%
\expandafter\ifx\csname l@#1\endcsname\relax
\typeout{** WARNING: IEEEtran.bst: No hyphenation pattern has been}%
\typeout{** loaded for the language `#1'. Using the pattern for}%
\typeout{** the default language instead.}%
\else
\language=\csname l@#1\endcsname
\fi
#2}}
\providecommand{\BIBdecl}{\relax}
\BIBdecl

\bibitem{lecun2015deep}
Y.~LeCun, Y.~Bengio, and G.~Hinton, ``Deep learning,'' \emph{Nature}, vol. 521,
  no. 7553, p. 436, 2015.

\bibitem{krizhevsky2012imagenet}
A.~Krizhevsky, I.~Sutskever, and G.~E. Hinton, ``Imagenet classification with
  deep convolutional neural networks,'' in \emph{Advances in Neural Information
  Processing Systems (NIPS)}, 2012, pp. 1097--1105.

\bibitem{russakovsky2015imagenet}
O.~Russakovsky, J.~Deng, H.~Su, J.~Krause, S.~Satheesh, S.~Ma, Z.~Huang,
  A.~Karpathy, A.~Khosla, M.~Bernstein \emph{et~al.}, ``Imagenet large scale
  visual recognition challenge,'' \emph{International Journal of Computer
  Vision}, vol. 115, no.~3, pp. 211--252, 2015.

\bibitem{canziani2016analysis}
A.~Canziani, A.~Paszke, and E.~Culurciello, ``An analysis of deep neural
  network models for practical applications,'' \emph{arXiv preprint
  arXiv:1605.07678}, 2016.

\bibitem{huang2017speed}
J.~Huang, V.~Rathod, C.~Sun, M.~Zhu, A.~Korattikara, A.~Fathi, I.~Fischer,
  Z.~Wojna, Y.~Song, S.~Guadarrama \emph{et~al.}, ``Speed/accuracy trade-offs
  for modern convolutional object detectors,'' in \emph{Proceedings of the IEEE
  Conference on Computer Vision and Pattern Recognition}, 2017, pp. 7310--7311.

\bibitem{paszke2017automatic}
A.~Paszke, S.~Gross, S.~Chintala, G.~Chanan, E.~Yang, Z.~DeVito, Z.~Lin,
  A.~Desmaison, L.~Antiga, and A.~Lerer, ``Automatic differentiation in
  pytorch,'' 2017.

\bibitem{github_celona}
S.~Bianco, R.~Cad\`ene, L.~Celona, and P.~Napoletano, ``Paper github
  repository,'' \url{https://github.com/CeLuigi/models-comparison.pytorch},
  2018 (accessed September 25, 2018).

\bibitem{simonyan2014very}
K.~Simonyan and A.~Zisserman, ``Very deep convolutional networks for
  large-scale image recognition,'' \emph{arXiv preprint arXiv:1409.1556}, 2014.

\bibitem{ioffe2015batch}
S.~Ioffe and C.~Szegedy, ``Batch normalization: Accelerating deep network
  training by reducing internal covariate shift,'' in \emph{International
  Conference on Machine Learning (ICML)}, 2015, pp. 448--456.

\bibitem{Szegedy_2015_CVPR}
C.~Szegedy, W.~Liu, Y.~Jia, P.~Sermanet, S.~Reed, D.~Anguelov, D.~Erhan,
  V.~Vanhoucke, and A.~Rabinovich, ``Going deeper with convolutions,'' in
  \emph{Conference on Computer Vision and Pattern Recognition (CVPR)}.\hskip
  1em plus 0.5em minus 0.4em\relax IEEE, 2015.

\bibitem{iandola2016squeezenet}
F.~N. Iandola, S.~Han, M.~W. Moskewicz, K.~Ashraf, W.~J. Dally, and K.~Keutzer,
  ``Squeezenet: Alexnet-level accuracy with 50x fewer parameters and< 0.5 mb
  model size,'' \emph{arXiv preprint arXiv:1602.07360}, 2016.

\bibitem{he2016deep}
K.~He, X.~Zhang, S.~Ren, and J.~Sun, ``Deep residual learning for image
  recognition,'' in \emph{Conference on Computer Vision and Pattern Recognition
  (CVPR)}.\hskip 1em plus 0.5em minus 0.4em\relax IEEE, 2016, pp. 770--778.

\bibitem{szegedy2016rethinking}
C.~Szegedy, V.~Vanhoucke, S.~Ioffe, J.~Shlens, and Z.~Wojna, ``Rethinking the
  inception architecture for computer vision,'' in \emph{Conference on Computer
  Vision and Pattern Recognition (CVPR)}.\hskip 1em plus 0.5em minus
  0.4em\relax IEEE, 2016, pp. 2818--2826.

\bibitem{szegedy2016inc}
C.~Szegedy, S.~Ioffe, V.~Vanhoucke, and A.~A. Alemi, ``Inception-v4,
  inception-resnet and the impact of residual connections on learning,'' in
  \emph{International Conference Learning Representations (ICLR) Workshop},
  2016.

\bibitem{huang2017densely}
G.~Huang, Z.~Liu, K.~Q. Weinberger, and L.~van~der Maaten, ``Densely connected
  convolutional networks,'' in \emph{Conference on Computer Vision and Pattern
  Recognition (CVPR)}, vol.~1, no.~2.\hskip 1em plus 0.5em minus 0.4em\relax
  IEEE, 2017, p.~3.

\bibitem{xie2017aggregated}
S.~Xie, R.~Girshick, P.~Doll{\'a}r, Z.~Tu, and K.~He, ``Aggregated residual
  transformations for deep neural networks,'' in \emph{Conference on Computer
  Vision and Pattern Recognition (CVPR)}.\hskip 1em plus 0.5em minus
  0.4em\relax IEEE, 2017, pp. 5987--5995.

\bibitem{Chollet_2017_CVPR}
F.~Chollet, ``Xception: Deep learning with depthwise separable convolutions,''
  in \emph{Conference on Computer Vision and Pattern Recognition (CVPR)}.\hskip
  1em plus 0.5em minus 0.4em\relax IEEE, 2017.

\bibitem{chen2017dual}
Y.~Chen, J.~Li, H.~Xiao, X.~Jin, S.~Yan, and J.~Feng, ``Dual path networks,''
  in \emph{Advances in Neural Information Processing Systems (NIPS)}, 2017, pp.
  4467--4475.

\bibitem{Hu_2018_CVPR}
J.~Hu, L.~Shen, and G.~Sun, ``Squeeze-and-excitation networks,'' in
  \emph{Conference on Computer Vision and Pattern Recognition (CVPR)}.\hskip
  1em plus 0.5em minus 0.4em\relax IEEE, 2018.

\bibitem{Zoph_2018_CVPR}
B.~Zoph, V.~Vasudevan, J.~Shlens, and Q.~V. Le, ``Learning transferable
  architectures for scalable image recognition,'' in \emph{Conference on
  Computer Vision and Pattern Recognition (CVPR)}.\hskip 1em plus 0.5em minus
  0.4em\relax IEEE, 2018.

\bibitem{howard2017mobilenets}
A.~G. Howard, M.~Zhu, B.~Chen, D.~Kalenichenko, W.~Wang, T.~Weyand,
  M.~Andreetto, and H.~Adam, ``Mobilenets: Efficient convolutional neural
  networks for mobile vision applications,'' \emph{arXiv preprint
  arXiv:1704.04861}, 2017.

\bibitem{sandler2018mobilenetv2}
M.~Sandler, A.~Howard, M.~Zhu, A.~Zhmoginov, and L.-C. Chen, ``Mobilenetv2:
  Inverted residuals and linear bottlenecks,'' in \emph{Conference on Computer
  Vision and Pattern Recognition (CVPR)}.\hskip 1em plus 0.5em minus
  0.4em\relax IEEE, 2018, pp. 4510--4520.

\bibitem{Zhang_2018_CVPR}
X.~Zhang, X.~Zhou, M.~Lin, and J.~Sun, ``Shufflenet: An extremely efficient
  convolutional neural network for mobile devices,'' in \emph{Conference on
  Computer Vision and Pattern Recognition (CVPR)}.\hskip 1em plus 0.5em minus
  0.4em\relax IEEE, 2018.

\bibitem{han2015deep}
S.~Han, H.~Mao, and W.~J. Dally, ``Deep compression: Compressing deep neural
  networks with pruning, trained quantization and huffman coding,'' \emph{arXiv
  preprint arXiv:1510.00149}, 2015.

\end{thebibliography}

\begin{IEEEbiography}[{\includegraphics[width=1in,height=1.25in,clip,keepaspectratio]{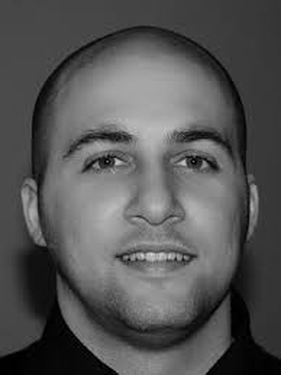}}]
{Simone Bianco} is assistant professor of Computer Science at Department of Informatics, Systems and Communication of the University of Milano-Bicocca, Italy. He obtained the PhD in Computer Science from the University of Milano-Bicocca, in 2010.
He obtained the BSc and the MSc degree in Mathematics from the University of Milano-Bicocca, respectively in 2003 and 2006. His current research interests include computer vision, machine learning, optimization algorithms, and color imaging.
\end{IEEEbiography}

\begin{IEEEbiography}[{\includegraphics[width=1in,height=1.25in,clip,keepaspectratio]{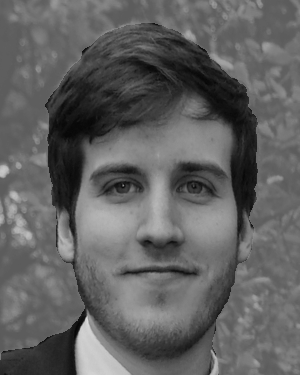}}]{Remi Cadene} is a PhD student and teaching assistant at LIP6 (Computer Science laboratory) of Sorbonne University, France. In 2016, he received a MSc degree in Computer Science at Sorbonne University. His primary research interests are in the fields of Machine Learning, Computer Vision and Natural Language Processing. He is currently focusing on Neural Networks, Multimodal Learning and Weakly Supervised Learning.
\end{IEEEbiography}

\begin{IEEEbiography}[{\includegraphics[width=1in,height=1.25in,clip,keepaspectratio]{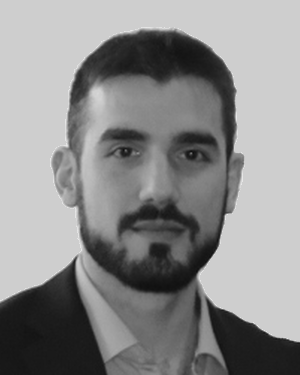}}]{Luigi Celona} is currently a postdoctoral fellow at DISCo (Department of Informatics, Systems and Communication) of the University of Milano-Bicocca, Italy. In 2018 and 2014, he obtained respectively the PhD and the MSc degree in Computer Science at DISCo. In 2011, he obtained the BSc degree in Computer Science from the University of Messina. His current research interests focus on image analysis and classification, machine learning and face analysis.
\end{IEEEbiography}

\begin{IEEEbiography}[{\includegraphics[width=1in,height=1.25in,clip,keepaspectratio]{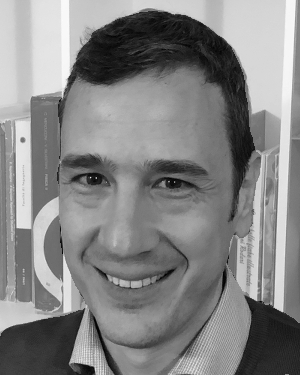}}]{Paolo Napoletano} is assistant professor of Computer Science at Department of Informatics, Systems and Communication of the University of Milano-Bicocca. In 2007, he received a PhD in Information Engineering from the University of Salerno. In 2003, he received a Master's degree in Telecommunications Engineering from the University of Naples Federico II. His current research interests include machine learning for multi-modal data classification and understanding.
\end{IEEEbiography}

\EOD

\end{document}